\pdfoutput=1
\documentclass[11pt]{article}

\usepackage[final]{acl}

\usepackage{times}
\usepackage{latexsym}

\usepackage[T1]{fontenc}
\usepackage[utf8]{inputenc}

\usepackage{microtype}

\usepackage{inconsolata}

\usepackage{graphicx}
\usepackage[inline]{enumitem}
\usepackage{subcaption}
\usepackage{booktabs} 
\usepackage{multirow}
\usepackage{float} % Include the float package in your preamble

\usepackage{amsmath}
\usepackage{amssymb}
\newcommand{\interalia}[1]{\citep[\emph{inter alia}]{#1}}

\usepackage{algorithm,algorithmicx}
\usepackage[noend]{algpseudocode}

\algnewcommand\algorithmicinput{\textbf{Input:}}
\algnewcommand\Input{\item[\algorithmicinput]}
\algnewcommand\algorithmicoutput{\textbf{Output:}}
\algnewcommand\Output{\item[\algorithmicoutput]}
\DeclareMathOperator*{\argmax}{argmax}
\usepackage{pgfplots}
\usepackage{filecontents}
\usepackage{filecontents}
\usepackage{pgfplots}
\pgfplotsset{compat=newest}
\usepgflibrary{plotmarks}
\usepgflibrary{plotmarks} % LATEX and plain TEX and pure pgf
\usepgflibrary[plotmarks] % ConTEXt and pure pgf
\usetikzlibrary{plotmarks} % LATEX and plain TEX when using TikZ
\usetikzlibrary[plotmarks] % ConTEXt when using TikZ
\usepgfplotslibrary{fillbetween}
\usepgfplotslibrary{groupplots}
\usepgfplotslibrary{groupplots, colorbrewer} % Load additional libraries
\usepackage{tabularx} % Add this for tabularx environment
\usepackage{xcolor} % For text color
\newcommand{\blue}[1]{\textcolor{blue}{#1}}
\usepackage{subcaption}
\usepackage{caption}
\usepackage{graphicx}
\usepackage{tikz}
\usepackage{xcolor}
\definecolor{customblue}{HTML}{DAE8FC}
\definecolor{customgreen}{HTML}{D5E8D4}
\definecolor{customred}{HTML}{F8CECC}
\definecolor{custompurple}{HTML}{E1D5E7}

\usepackage{bbm}

\title{ALVIN: Active Learning Via INterpolation}

\author{
Michalis Korakakis$^{1,3}$ \ \ \ \ \ Andreas Vlachos$^{1}$ \ \ \ \ \ Adrian Weller$^{2,3}$  \\
 $^1$Department of Computer Science and Technology, University of Cambridge\\
 $^2$Department of Engineering, University of Cambridge\\
 $^3$The Alan Turing Institute \\
  {\texttt{\{mk2008,av308,aw665\}@cam.ac.uk}} \\
}

\begin{document}
\maketitle
\begin{abstract}
Active Learning aims to minimize annotation effort by selecting the most useful instances from a pool of unlabeled data. However, typical active learning methods overlook the presence of distinct example groups within a class, whose prevalence may vary, e.g., in occupation classification datasets certain demographics are disproportionately represented in specific classes. This oversight causes models to rely on shortcuts for predictions, i.e., spurious correlations between input attributes and labels occurring in well-represented groups. To address this issue, we propose Active Learning Via INterpolation~(ALVIN), which conducts intra-class interpolations between examples from under-represented and well-represented groups to create anchors, i.e., artificial points situated between the example groups in the representation space. By selecting instances close to the anchors for annotation, ALVIN identifies informative examples exposing the model to regions of the representation space that counteract the influence of shortcuts. Crucially, since the model considers these examples to be of high certainty, they are likely to be ignored by typical active learning methods. Experimental results on six datasets encompassing sentiment analysis, natural language inference, and paraphrase detection demonstrate that ALVIN outperforms state-of-the-art active learning methods in both in-distribution and out-of-distribution generalization.
\end{abstract}

\begin{figure}[th!]
    \centering
    \includegraphics[width=\linewidth]{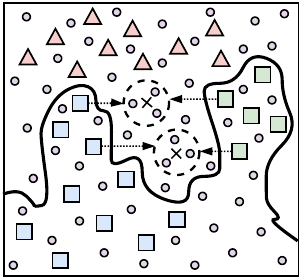}
    \caption{Illustration of ALVIN applied to a binary classification task. 
        \protect\raisebox{-0.4ex}{\protect\tikz[baseline=-0.4ex]\protect\draw[fill=customblue] (0,0) rectangle (0.75em,0.75em);} indicates well-represented, labeled examples in Class A, 
        \protect\raisebox{-0.4ex}{\protect\tikz[baseline=-0.4ex]\protect\draw[fill=customgreen] (0,0) rectangle (0.75em,0.75em);} indicates under-represented, labeled examples in Class A, 
        \protect\raisebox{-0.5ex}{\protect\tikz[baseline=-0.5ex]\protect\draw[fill=customred] (0,0) -- (0.75em,0) -- (0.375em,0.75em) -- cycle;} indicates labeled examples in Class B, 
        \protect\raisebox{+0.2ex}{\protect\tikz[baseline=-0.4ex]\protect\draw[fill=custompurple] (0,0) circle (0.22em);}~indicates unlabeled instances, and 
        \protect\raisebox{-0.3ex}{\protect\tikz[baseline=-0.3ex]\protect\draw[line width=0.6pt] (0,0) -- (0.6em,0.6em) (0,0.6em) -- (0.6em,0);} indicates the anchors created via intra-class interpolations between under-represented and well-represented examples. Unlike typical active learning methods, ALVIN prioritizes high-certainty instances that integrate representations from different example groups at varied proportions. This approach enables ALVIN to adjust the model's decision boundary and mitigate its reliance on shortcuts.
    }
    \label{fig:alvin}
\end{figure}

\section{Introduction}
Despite the remarkable zero-shot and few-shot learning capabilities of large language models~(LLMs) \interalia{DBLP:conf/nips/BrownMRSKDNSSAA20, DBLP:journals/jmlr/ChowdheryNDBMRBCSGSSTMRBTSPRDHPBAI23, DBLP:journals/corr/abs-2307-09288}, supervised fine-tuning remains a critical component of model development~\citep{DBLP:journals/corr/abs-2308-01825, mosbach-etal-2023-shot, DBLP:journals/corr/abs-2310-19462}. Collecting high-quality labeled data is, nonetheless, time-consuming and labor-intensive~\citep{DBLP:journals/corr/abs-2402-13446}. To address this annotation bottleneck, active learning~(AL) seeks to select the most useful instances from a pool of unlabeled data, thereby maximizing model performance subject to an annotation budget~\citep{Settles2009ActiveLL}.

However, datasets commonly used for model fine-tuning often contain shortcuts~\citep{gururangan-etal-2018-annotation, mccoy-etal-2019-right, wang-culotta-2020-identifying}, i.e., spurious correlations between input attributes and labels present in a large number of examples~\citep{DBLP:journals/natmi/GeirhosJMZBBW20}. For example, in occupation classification datasets, many examples exhibit patterns that incorrectly associate certain demographics, such as race and gender, with specific occupations~\citep{ DBLP:conf/www/BorkanDSTV19}. Consequently, models exploiting shortcuts achieve high performance on well-represented example groups, but fail on under-represented groups where shortcuts do not apply~\citep{tu-etal-2020-empirical}. This issue is particularly prominent in out-of-distribution settings, where under-represented groups can become more prevalent due to distribution shifts~\citep{DBLP:conf/icml/KohSMXZBHYPGLDS21}. By neglecting the presence of these distinct example groups in the training data, AL methods amplify the prevalence of well-represented groups, thereby exacerbating shortcut learning~\citep{DBLP:conf/cvpr/GudovskiyHYT20,deng-etal-2023-counterfactual}.

Motivated by these shortcomings, we introduce Active Learning Via INterpolation~(ALVIN). The key idea behind ALVIN is to leverage interpolations between example groups to explore the representation space. Specifically, we identify unlabeled instances for annotation by assessing their proximity to anchors, i.e., artificial points in the representation space created through intra-class interpolations between under-represented and well-represented examples. Intuitively, ALVIN selects informative instances with features distinct from those prevalent in well-represented groups, helping the model avoid reliance on shortcuts. Importantly, because these instances are deemed high certainty by the model, they are often overlooked by typical AL methods.

We conduct experiments on six datasets spanning sentiment analysis, natural language inference, and paraphrase detection. Our results demonstrate that ALVIN consistently improves out-of-distribution generalization compared to several state-of-the-art AL methods, across different dataset acquisition sizes, while also maintaining high in-distribution performance. 

We analyze ALVIN to gain deeper insights into its performance improvements. First, we examine the unlabeled examples identified by ALVIN, showcasing its ability to select diverse, high-certainty instances while avoiding outliers that could negatively impact performance. Next, through several ablation studies, we demonstrate the advantages of our interpolation strategy compared to other interpolation-based AL methods. Finally, we explore the impact of hyper-parameters on performance and assess the computational runtime required to select instances for annotation.

\begin{algorithm*}[]
\caption{Active~Learning~Via~INterpolation~(ALVIN)}
\label{algo:alvin}
\begin{algorithmic}[1]
    \Input Training dataset $\mathcal{L}$, unlabeled pool $\mathcal{U}$, model $f_\theta = \{f_{\text{enc}}, f_{\text{cls}}\}$, annotation batch size $b$, number of anchors $K$, shape parameter $\alpha$ of Beta distribution
  
    \State{$\mathcal{I} = \emptyset$}
    \State{$g_\text{min}, g_\text{maj}= \textsc{InferMinMaj}(f_{{\theta}}, \mathcal{L})$}~\label{algo:infer_examples}

    \For{$c \in \mathcal{C}$}

    \State{Sample $\mathcal{L}_c^{\text{min}}, \mathcal{L}_c^{\text{maj}}  \sim \mathcal{L}$} %\Comment{$\mathcal{L}_c^{\text{min}}$ and $\mathcal{L}_c^{\text{maj}}$ contain an equal number of instances}
    \label{algo:sample}

    \For{$(x_i, y_i) \in \mathcal{L}_c^{\text{min}}$}\label{algo:iterate_min}
        %\If{$(x_i, y_i) \in g_\text{min}$}
        \State{Sample $(x_j, y_j) \sim g_{c}^{\text{maj}}$}\label{algo:sample_maj}
        %\Else
        %    \State{Sample $(x_j, y_j) \sim g_{c}^{\text{min}}^$} 
        %\EndIf      
        \For {$k$ in $K$} \Comment{{generate multiple anchors}}\label{algo:vary_lambda}
            \State{Sample $\lambda \sim \text{Beta}(\alpha, \alpha)$ }
            \State{${a_{i, j}^{k}}=\lambda f_\text{enc}(x_{i}) + (1-\lambda) f_\text{enc}(x_j)$} \;\label{algo:interpolate}
            \State{$\mathcal{I} \gets \mathcal{I} \cup  \underset{x \in \mathcal{U}} {\text{Top-k}} \ \mathrm{KNN}\big({a_{i, j}^{k}}, \mathcal{U})$} \Comment{select k nearest neighbors of anchor from $\mathcal{U}$}\label{algo:knn}
        \EndFor
    \EndFor
\EndFor
\State{$B = \underset{x \in \mathcal{I}}\argmax - \sum \limits_{i=1}^{C} f_\text{cls}(f_\text{enc}(x))_{i} \log{f_\text{cls}(f_\text{enc}(x))_{i}}, |B|=b$} \Comment{{select top-b instances via uncertainty}}\;\label{algo:uncertainty}
\end{algorithmic}
\end{algorithm*}

\section{Active Learning Via INterpolation}

\subsection{Preliminaries}
We consider the typical pool-based active learning~(AL) scenario~\citep{DBLP:conf/sigir/LewisG94}, in which an initial set of labeled instances $\mathcal{L} = \{(x_i,y_i)\}_{i=1}^N$, where $x_{i} \in \mathcal{X}$ is the input and $y_{i} \in \{1,2,\dots,C\}$ is the corresponding label, along with a pool of unlabeled instances $\mathcal{U} = \{x_j\}_{j=1}^M$, where $N \ll M$. In each AL round, we query an annotation batch $B$ comprised of $b$ instances from $\mathcal{U}$ to be annotated and added to $\mathcal{L}$. Then $\mathcal{L}$ is used to train a model $f_\theta: \mathcal{X} \rightarrow \mathcal{Y}$ parameterized by $\theta$. The model $f_\theta$ consists of an encoder $f_\text{enc}: \mathcal{X} \rightarrow \mathcal{Z}$ mapping an input $x_{i}$ to a representation $z_{i}$, and a classifier $f_\text{cls}: \mathcal{Z} \rightarrow \mathcal{Y}$ which outputs a softmax probability over the labels based on $z_{i}$. The AL process continues until the annotation budget is exhausted or a satisfactory model performance level is reached.

Following~\citet{DBLP:journals/corr/abs-1911-08731}, we further assume that the training dataset contains distinct groups of instances within some classes. Some of these groups are well-represented and strongly associated with labels, e.g., high word overlap and ``entailment'' in natural language inference~(NLI) datasets~\citep{mccoy-etal-2019-right}, while others are under-represented, e.g., negation in the hypothesis and ``entailment''~\citep{gururangan-etal-2018-annotation}. We refer to the instances belonging to the well-represented groups associated with a particular class as majority instances $g_{\text{maj}}$ of said class, and the rest as minority instances $g_{\text{min}}$.\footnote{Note that some instances can be majority for a particular class, and other instances exhibit the same patterns can be minority for a different class e.g., NLI instances containing negation in the hypothesis are majority for the ``contradiction'' class, but minority for the ``entailment'' class.}

Models often rely on shortcuts found in majority instances to make predictions~\citep{DBLP:journals/corr/abs-2308-12553}, a dependency that becomes problematic when distribution shifts at test time increase the prevalence of minority examples, resulting in poor out-of-distribution generalization~\citep{DBLP:conf/icml/KohSMXZBHYPGLDS21}. This issue is further exacerbated in AL, where typical methods like uncertainty sampling~\citep{DBLP:conf/sigir/LewisG94}, select repetitive high uncertainty majority instances~\citep{deng-etal-2023-counterfactual}. To counter shortcut learning, it is crucial for the model to be exposed to instances whose patterns deviate from those prevalent in majority examples~\citep{korakakis-vlachos-2023-improving}.

\subsection{Algorithm}
We hypothesize that the properties of the representation space are crucial for identifying unlabeled instances capable of mitigating shortcut learning. Specifically, the reliance on shortcuts for predictions creates a spurious decision boundary, incorrectly separating minority and majority examples within the same class. Thus, our goal is to select informative instances that will prompt the model to adjust its decision boundary, thereby correcting its reliance on shortcut features. To achieve this, ALVIN employs intra-class interpolations between minority and majority instances to create anchors. These anchors facilitate the exploration of diverse feature combinations within the representation space, enabling the identification of unlabeled instances that integrate representations from different example groups at varied proportions. 
However, because these instances exhibit high certainty, they are typically overlooked by existing AL methods, e.g., a model will confidently label an ``entailment'' instance with negation in NLI as ``contradiction.'' The overall procedure of ALVIN is detailed in Algorithm~\ref{algo:alvin} for an AL round.

\paragraph{Inferring Minority/Majority Examples}
At the beginning of each AL round, we first identify the minority and majority examples within each class in the training dataset~(line~\ref{algo:infer_examples}). We are motivated by the observation that the existence of shortcuts within the majority examples causes a discrepancy in training dynamics, leading the model to fit majority examples faster than minority ones, and resulting in a spurious decision boundary~\citep{DBLP:conf/nips/ShahTR0N20, tu-etal-2020-empirical, DBLP:conf/nips/PezeshkiKBCPL21}. Thus, we infer the example groups by monitoring the frequency with which the model incorrectly predicts an example~\citep{DBLP:conf/iclr/TonevaSCTBG19, swayamdipta-etal-2020-dataset, yaghoobzadeh-etal-2021-increasing}. Specifically, we classify an example \(x_i\) as minority~\begin{enumerate*}[label=(\arabic*), before=\unskip{ if }, itemjoin={{, or }}]

\item the model's predictions switch between correct to incorrect at least once during training, i.e., \(\text{acc}_{x_i}^t > \text{acc}_{x_{i}}^{t+1}\), where \(\text{acc}_{x_{i}}^t = \mathbbm{1}_{\hat{y}_i^t=y_i}\) indicates that the example \(x_i\) is correctly classified at time step \(t\)

\item the example is consistently misclassified by the model throughout training, i.e., $
\forall t \in \{1, 2, \dots, T\}, \quad \text{acc}_{x_i}^t = {0}
$ where \( T \) is the total number of training epochs.
\end{enumerate*} Conversely, all other examples that do not meet these criteria are classified as majority examples.

\paragraph{Anchor Creation}
After identifying the minority and majority examples within each class, we then proceed to create anchors to explore the representation space between these example groups. In particular, for each class $c$ in $\mathcal{C}$, we initially sample $\mathcal{L}^{\text{min}}_{c}$ and $\mathcal{L}^{\text{maj}}_{c}$~(line~\ref{algo:sample}), where $|\mathcal{L}^{\text{min}}_{c}| = |\mathcal{L}^{\text{maj}}_{c}| \ll N$. Next, for every minority instance in $\mathcal{L}^{\text{min}}_c$~(line~\ref{algo:iterate_min}) we randomly sample a majority instance from $\mathcal{L}^{\text{maj}}_c$~(line~\ref{algo:sample_maj}), and interpolate their representations to create the anchor $a_{i, j}$~(line~\ref{algo:interpolate}):

\begin{align}
    a_{i, j}= \lambda f_\text{enc}(x_{i}) + (1 - \lambda) f_\text{enc}(x_j),
\label{eq:interpolation}
\end{align}

\noindent where the interpolation ratio $\lambda \in [0,1]$ is sampled from a Beta distribution $\text{Beta}(\alpha, \alpha)$. By adjusting the parameter $\alpha$ of this distribution, we can control where the anchors lie in the representation space relative to minority or majority instances. Intuitively, when $\lambda$ is closer to 0, the anchor $a_{i, j}$ is predominantly influenced by the minority instance $x_{i}$; conversely, as $\lambda$ approaches 1, $a_{i, j}$ increasingly resembles the representation of majority instance~$x_{j}$.

We generate $K$ anchors for each minority-majority pair~(line~\ref{algo:vary_lambda}). This process enables us to create anchors that incorporate varied feature combinations, thus allowing for a comprehensive exploration of the representation space between minority and majority examples.

\paragraph{Example Selection}
After constructing the anchors, we use K-Nearest-Neighbors~(KNN) to identify similar unlabeled examples $x_u \in \mathcal{U}$ to an anchor in the representation space~(line~\ref{algo:knn}).\footnote{Our distance metric is the Euclidean distance.} We repeat this process for each anchor across all classes. Finally, we select for annotation the top-$b$ unlabeled instances with the highest uncertainty~\citep{DBLP:conf/sigir/LewisG94}~(line~\ref{algo:uncertainty}). This approach maintains the advantages of uncertainty-based instance selection, while counteracting its tendency to facilitate shortcut learning by selecting a subset of unlabeled instances that mitigate this phenomenon.%, i.e., XXXXXXXX.

\section{Experimental Setup}
\paragraph{Datasets} We conduct experiments on six datasets across sentiment analysis, natural language inference, and paraphrase detection. In line with previous works in AL~\citep{yuan-etal-2020-cold, margatina-etal-2021-active, deng-etal-2023-counterfactual}, we use SA~\citep{DBLP:conf/iclr/KaushikHL20}, NLI~\citep{DBLP:conf/iclr/KaushikHL20}, ANLI~\citep{nie-etal-2020-adversarial}, SST-2~\citep{socher-etal-2013-recursive}, IMDB~\citep{maas-etal-2011-learning}, and QQP~\citep{Chen2017QuoraQP}. To assess out-of-distribution~(OOD) generalization we use SemEval-2017 Task 4~\citep{rosenthal-etal-2017-semeval} for SA, ANLI for NLI, and NLI for ANLI, IMDB for SST-2, SST-2 for IMDB, and TwitterPPDB~\citep{lan-etal-2017-continuously} for QQP. Validation and test splits are used as described in~\citet{margatina-etal-2021-active} for IMDB, SST-2, and QQP, and~\citet{deng-etal-2023-counterfactual} for SA, ANLI, and NLI.

\paragraph{Comparisons} We compare ALVIN with several baseline and state-of-the-art AL methods: 
\begin{itemize}[leftmargin=12pt,noitemsep,topsep=0pt,parsep=2pt]

    \item \textbf{Random} samples instances uniformly at random. 
   
    \item \textbf{Uncertainty}~\citep{DBLP:conf/sigir/LewisG94} acquires annotations for unlabeled instances with the highest predictive entropy according to the model.

    \item \textbf{Batch Active learning by Diverse Gradient Embeddings~(BADGE)}~\citep{DBLP:conf/iclr/AshZK0A20} selects unlabeled instances by applying the K-means++~\citep{DBLP:conf/soda/ArthurV07} clustering algorithm on the gradients of the predicted class with respect to the model's last layer.

    \item \textbf{BERT-KM}~\citep{yuan-etal-2020-cold} clusters unlabeled instances within the representation space of a BERT~\citep{devlin-etal-2019-bert} model using k-means, then selects for annotation those instances that are closest to the center of each cluster.

    \item \textbf{Contrastive Active Learning~(CAL)}~\citep{margatina-etal-2021-active} selects unlabeled instances that, according to the model, diverge maximally from their nearest labeled neighbors.

    \item \textbf{Active Learning by Feature Mixing~(ALFA-Mix)}~\citep{DBLP:conf/cvpr/ParvanehATHHS22} conducts interpolations between unlabeled instances and anchors, i.e., the average embeddings of the labeled examples for each class, and then selects unlabeled instances whose interpolations have different predictions compared to the anchors.

\end{itemize}

\begin{table*}[th!]
\centering
\setlength{\tabcolsep}{0.15em}
\resizebox{\textwidth}{!}{% Adjusted to use full linewidth, be cautious with 1.0 as it might just touch the margins
\begin{tabular}{cc|cccccccccccccccc}
\toprule
\multirow{2}{*}{\textbf{Data}} & \multirow{2}{*}{\textbf{Acq. Data}} &
\multicolumn{2}{c}{\textbf{Random}} & \multicolumn{2}{c}{\textbf{Uncertainty}} &
\multicolumn{2}{c}{\textbf{BADGE}} & \multicolumn{2}{c}{\textbf{BERT-KM}} &
\multicolumn{2}{c}{\textbf{CAL}} & \multicolumn{2}{c}{\textbf{ALFA-Mix}} &
\multicolumn{2}{c}{\textbf{ALVIN}} \\
 & & \textbf{ID} & \textbf{OOD} & \textbf{ID} & \textbf{OOD} & \textbf{ID} & \textbf{OOD} & \textbf{ID} & \textbf{OOD} & \textbf{ID} & \textbf{OOD} & \textbf{ID} & \textbf{OOD} & \textbf{ID} & \textbf{OOD} \\
\midrule
\multirow{3}{*}{\textbf{SA}}  
&       1\%                                 & \textbf{78.9}\textsubscript{\tiny{$\pm$0.2}}  & 59.4\textsubscript{\tiny{$\pm$1.8}}    & 69.7\textsubscript{\tiny{$\pm$0.2}} &  57.9\textsubscript{\tiny{$\pm$2.7}}        & 74.6\textsubscript{\tiny{$\pm$0.5}} & 56.2\textsubscript{\tiny{$\pm$1.9}}   &  66.4\textsubscript{\tiny{$\pm$0.4}}   &  {60.5}\textsubscript{\tiny{$\pm$3.5}}   & 72.4\textsubscript{\tiny{$\pm$0.2}}  &  57.8\textsubscript{\tiny{$\pm$3.5}}        & 73.9\textsubscript{\tiny{$\pm$0.5}}    & 58.0\textsubscript{\tiny{$\pm$2.5}} & {77.9}\textsubscript{\tiny{$\pm$0.7}} & \underline{61.5}\textsubscript{\tiny{$\pm$0.5}} \\

&   5\%         & 86.9\textsubscript{\tiny{$\pm$0.1}} &   73.9\textsubscript{\tiny{$\pm$2.2}}               & \textbf{90.8}\textsubscript{\tiny{$\pm$0.3}} & 74.4\textsubscript{\tiny{$\pm$3.2}}               &  88.9\textsubscript{\tiny{$\pm$0.6}} & 79.7\textsubscript{\tiny{$\pm$2.1}}              &  90.2\textsubscript{\tiny{$\pm$0.3}}  & 75.6\textsubscript{\tiny{$\pm$3.4}}           & 89.4\textsubscript{\tiny{$\pm$0.3}}   & 79.3\textsubscript{\tiny{$\pm$3.1}}       & 89.7\textsubscript{\tiny{$\pm$0.9}}    & {79.8}\textsubscript{\tiny{$\pm$3.2}} & \textbf{90.8}\textsubscript{\tiny{$\pm$1.0}} & \underline{82.2}\textsubscript{\tiny{$\pm$1.2}} \\

&    10\%                                   & 88.3\textsubscript{\tiny{$\pm$0.2}}  & {81.1}\textsubscript{\tiny{$\pm$1.9}}                & {91.1}\textsubscript{\tiny{$\pm$0.3}} &  78.2\textsubscript{\tiny{$\pm$3.4}}               & 90.2\textsubscript{\tiny{$\pm$0.4}} & 78.3\textsubscript{\tiny{$\pm$1.8}}               &  88.3\textsubscript{\tiny{$\pm$0.5}}   & 75.9\textsubscript{\tiny{$\pm$2.8}}           & 90.5\textsubscript{\tiny{$\pm$0.2}}  &  73.0\textsubscript{\tiny{$\pm$2.3}}      & 90.5\textsubscript{\tiny{$\pm$0.7}}    & 78.4\textsubscript{\tiny{$\pm$2.9}} & \textbf{91.8}\textsubscript{\tiny{$\pm$1.3}} & \underline{84.1}\textsubscript{\tiny{$\pm$0.9}} \\

\midrule

\multirow{3}{*}{\textbf{NLI}}  
&    1\%                                & \textbf{44.7}\textsubscript{\tiny{$\pm$0.6}}     &  34.2\textsubscript{\tiny{$\pm$0.9}}               &  41.2\textsubscript{\tiny{$\pm$1.2}}    &  33.2\textsubscript{\tiny{$\pm$1.7}}          &  41.3\textsubscript{\tiny{$\pm$1.3}}    &   33.8\textsubscript{\tiny{$\pm$1.2}}     & 42.4 \textsubscript{\tiny{$\pm$1.6}}     &  34.7\textsubscript{\tiny{$\pm$1.0}}      & 43.3\textsubscript{\tiny{$\pm$0.4}}     &  {35.3}\textsubscript{\tiny{$\pm$0.7}}      & 42.8\textsubscript{\tiny{$\pm$1.4}}    & 34.6\textsubscript{\tiny{$\pm$2.2}} & {43.4}\textsubscript{\tiny{$\pm$0.8}} & \underline{35.7}\textsubscript{\tiny{$\pm$1.5}}\\
&    5\%                                & 67.1\textsubscript{\tiny{$\pm$0.9}}     &  35.8\textsubscript{\tiny{$\pm$1.1}}               &63.9\textsubscript{\tiny{$\pm$1.4}}     &   35.7\textsubscript{\tiny{$\pm$1.9}}          & 63.7\textsubscript{\tiny{$\pm$1.2}}     &   35.0\textsubscript{\tiny{$\pm$1.4}}     & 65.8\textsubscript{\tiny{$\pm$1.8}}    &  34.6\textsubscript{\tiny{$\pm$1.2}}      & {67.8}\textsubscript{\tiny{$\pm$0.4}}     &  36.0\textsubscript{\tiny{$\pm$1.2}}      & {67.8}\textsubscript{\tiny{$\pm$1.7}}    & {36.3}\textsubscript{\tiny{$\pm$1.9}} & \textbf{69.7}\textsubscript{\tiny{$\pm$1.1}} & \underline{38.9}\textsubscript{\tiny{$\pm$0.7}}\\
&    10\%                               & 72.9\textsubscript{\tiny{$\pm$0.6}}     &  37.9\textsubscript{\tiny{$\pm$0.8}}               & {76.2}\textsubscript{\tiny{$\pm$1.0}}     &    37.9\textsubscript{\tiny{$\pm$1.3}}         & 76.1\textsubscript{\tiny{$\pm$1.4}}     &   37.0\textsubscript{\tiny{$\pm$1.4}}     & 73.1\textsubscript{\tiny{$\pm$1.5}}      & 37.6\textsubscript{\tiny{$\pm$1.2}}        & 77.6\textsubscript{\tiny{$\pm$0.6}}     &  {39.9}\textsubscript{\tiny{$\pm$0.8}}      & {77.7}\textsubscript{\tiny{$\pm$2.1}}    & {40.1}\textsubscript{\tiny{$\pm$3.1}} & \textbf{78.1}\textsubscript{\tiny{$\pm$1.1}} & \underline{42.9}\textsubscript{\tiny{$\pm$1.5}} \\
\midrule

\multirow{3}{*}{\textbf{ANLI}}  
&    1\%                                &  34.1\textsubscript{\tiny{$\pm$0.4}}       & 33.1\textsubscript{\tiny{$\pm$1.3}}       & 33.1\textsubscript{\tiny{$\pm$1.4}}        &  {34.1}\textsubscript{\tiny{$\pm$2.4}}         & \textbf{34.8}\textsubscript{\tiny{$\pm$1.4}}      &   32.8\textsubscript{\tiny{$\pm$1.7}}      & 33.4 \textsubscript{\tiny{$\pm$1.2}}        & 33.3\textsubscript{\tiny{$\pm$1.3}}         &  33.0\textsubscript{\tiny{$\pm$1.1}}    &   \underline{34.5}\textsubscript{\tiny{$\pm$2.4}}      & 33.3\textsubscript{\tiny{$\pm$1.2}}    & 33.7\textsubscript{\tiny{$\pm$1.7}} & {34.2}\textsubscript{\tiny{$\pm$0.5}} & 33.8\textsubscript{\tiny{$\pm$0.9}}\\
&    5\%                                & 36.4\textsubscript{\tiny{$\pm$0.3}}        &  35.1\textsubscript{\tiny{$\pm$0.9}}        & {37.3}\textsubscript{\tiny{$\pm$1.4}}        &  {35.9}\textsubscript{\tiny{$\pm$1.9}}        &  {37.3}\textsubscript{\tiny{$\pm$1.5}}     &   34.6\textsubscript{\tiny{$\pm$1.7}}     & 36.6\textsubscript{\tiny{$\pm$1.2}}        &  32.4\textsubscript{\tiny{$\pm$1.2}}       & 36.2\textsubscript{\tiny{$\pm$1.3}}      & 34.1\textsubscript{\tiny{$\pm$1.9}}       &  \textbf{37.8}\textsubscript{\tiny{$\pm$1.8}}    & 34.7\textsubscript{\tiny{$\pm$2.4}} & {37.4}\textsubscript{\tiny{$\pm$0.9}} & \underline{37.9}\textsubscript{\tiny{$\pm$0.6}}\\
&    10\%                               &  38.9\textsubscript{\tiny{$\pm$0.4}}       &  33.5\textsubscript{\tiny{$\pm$1.2}}        & {39.9}\textsubscript{\tiny{$\pm$1.7}}        &  35.9\textsubscript{\tiny{$\pm$2.7}}         & {41.0}\textsubscript{\tiny{$\pm$1.2}}       &  36.0\textsubscript{\tiny{$\pm$1.5}}     & 40.1\textsubscript{\tiny{$\pm$1.3}}      & 31.5\textsubscript{\tiny{$\pm$1.1}}         & 38.3\textsubscript{\tiny{$\pm$1.2}}     &  35.2\textsubscript{\tiny{$\pm$2.2}}       & 38.3\textsubscript{\tiny{$\pm$1.8}}    & {36.1}\textsubscript{\tiny{$\pm$2.3}} & \textbf{42.6}\textsubscript{\tiny{$\pm$1.0}} & \underline{39.2}\textsubscript{\tiny{$\pm$1.3}}\\

\midrule

\multirow{3}{*}{\textbf{SST-2}}  
&    1\%        &  84.0\textsubscript{\tiny{$\pm$0.5}}      &     69.3\textsubscript{\tiny{$\pm$0.7}}    &  84.6\textsubscript{\tiny{$\pm$0.8}}        &    68.6\textsubscript{\tiny{$\pm$1.5}}      &  84.6\textsubscript{\tiny{$\pm$0.6}}       &    68.6\textsubscript{\tiny{$\pm$1.1}}    & 84.7\textsubscript{\tiny{$\pm$0.9}}       &     68.6\textsubscript{\tiny{$\pm$1.4}}     &  85.0\textsubscript{\tiny{$\pm$0.6}}     &   69.8\textsubscript{\tiny{$\pm$0.7}}      & {85.9}\textsubscript{\tiny{$\pm$0.7}}    &  {70.6}\textsubscript{\tiny{$\pm$0.6}} & \textbf{86.8}\textsubscript{\tiny{$\pm$0.3}} & \underline{71.9}\textsubscript{\tiny{$\pm$0.9}}\\
&    5\%        & 86.4\textsubscript{\tiny{$\pm$0.7}}        &    71.8\textsubscript{\tiny{$\pm$0.6}}     &  87.9\textsubscript{\tiny{$\pm$0.7}}      &      70.3\textsubscript{\tiny{$\pm$1.3}}    &   87.3\textsubscript{\tiny{$\pm$0.8}}     &   70.9\textsubscript{\tiny{$\pm$1.2}}    & {88.8}\textsubscript{\tiny{$\pm$0.5}}        &     70.9\textsubscript{\tiny{$\pm$0.7}}    &  87.7\textsubscript{\tiny{$\pm$0.6}}     &    73.6\textsubscript{\tiny{$\pm$1.2}}    & 87.9\textsubscript{\tiny{$\pm$0.6}}  &  {74.2}\textsubscript{\tiny{$\pm$0.8}} & \textbf{90.0}\textsubscript{\tiny{$\pm$0.3}} & \underline{77.6}\textsubscript{\tiny{$\pm$0.9}}\\
&    10\%       &   88.1\textsubscript{\tiny{$\pm$0.7}}       &  73.1\textsubscript{\tiny{$\pm$0.9}}       &  89.3\textsubscript{\tiny{$\pm$0.5}}       &     72.1\textsubscript{\tiny{$\pm$1.1}}     &  88.7\textsubscript{\tiny{$\pm$0.6}}       &   71.2\textsubscript{\tiny{$\pm$1.4}}    & 89.3\textsubscript{\tiny{$\pm$1.8}}       &   71.4\textsubscript{\tiny{$\pm$0.9}}       &  {89.4}\textsubscript{\tiny{$\pm$0.4}}      &    75.4\textsubscript{\tiny{$\pm$0.8}}   & 89.0\textsubscript{\tiny{$\pm$0.5}}  & {76.3}\textsubscript{\tiny{$\pm$1.4}}  & \textbf{90.1}\textsubscript{\tiny{$\pm$0.5}} & \underline{78.9}\textsubscript{\tiny{$\pm$0.8}}\\

\midrule

\multirow{3}{*}{\textbf{IMDB}}  
&    1\%        &  66.1\textsubscript{\tiny{$\pm$0.6}}        &    59.4\textsubscript{\tiny{$\pm$1.8}}     &  68.4\textsubscript{\tiny{$\pm$0.6}}        &    60.6\textsubscript{\tiny{$\pm$1.0}}      &  68.1\textsubscript{\tiny{$\pm$0.5}}     &   60.3\textsubscript{\tiny{$\pm$2.7}}    & 68.3\textsubscript{\tiny{$\pm$1.6}}       &     60.1\textsubscript{\tiny{$\pm$1.5}}     &   {73.7}\textsubscript{\tiny{$\pm$0.5}}    &   60.6\textsubscript{\tiny{$\pm$1.2}}      &  73.6\textsubscript{\tiny{$\pm$0.5}}    & {61.4}\textsubscript{\tiny{$\pm$1.8}}  & \textbf{74.2}\textsubscript{\tiny{$\pm$1.5}} & \underline{63.7}\textsubscript{\tiny{$\pm$0.6}}\\
&    5\%        & 84.4\textsubscript{\tiny{$\pm$0.7}}        &   77.3\textsubscript{\tiny{$\pm$1.6}}      &  84.8\textsubscript{\tiny{$\pm$0.6}}        &    {80.3}\textsubscript{\tiny{$\pm$0.9}}      &   84.6\textsubscript{\tiny{$\pm$0.5}}    &   79.6\textsubscript{\tiny{$\pm$3.3}}    & {84.8}\textsubscript{\tiny{$\pm$0.8}}        &     79.1\textsubscript{\tiny{$\pm$2.3}}    &  {84.9}\textsubscript{\tiny{$\pm$0.4}}      &       79.4\textsubscript{\tiny{$\pm$0.7}} &  84.5\textsubscript{\tiny{$\pm$0.5}}    & {80.3}\textsubscript{\tiny{$\pm$2.0}}  & \textbf{86.5}\textsubscript{\tiny{$\pm$1.2}} & \underline{84.0}\textsubscript{\tiny{$\pm$0.3}}\\
&    10\%       &  86.3\textsubscript{\tiny{$\pm$0.6}}        &   79.6\textsubscript{\tiny{$\pm$2.9}}     &  87.1\textsubscript{\tiny{$\pm$0.6}}         &     {82.4}\textsubscript{\tiny{$\pm$1.2}}     & 87.2\textsubscript{\tiny{$\pm$0.4}}       &   81.7\textsubscript{\tiny{$\pm$3.1}}    & {87.4}\textsubscript{\tiny{$\pm$1.5}}        &    81.2\textsubscript{\tiny{$\pm$1.5}}      &  {87.4}\textsubscript{\tiny{$\pm$0.5}}    &    81.3\textsubscript{\tiny{$\pm$0.6}}      &  {87.4}\textsubscript{\tiny{$\pm$0.6}}  &  82.2\textsubscript{\tiny{$\pm$2.1}} & \textbf{88.8}\textsubscript{\tiny{$\pm$0.9}} & \underline{84.8}\textsubscript{\tiny{$\pm$0.7}} \\

\midrule

\multirow{3}{*}{\textbf{QQP}}  
&    1\%        &  77.5\textsubscript{\tiny{$\pm$0.6}}       &    {71.3}\textsubscript{\tiny{$\pm$0.3}}     &  {78.6}\textsubscript{\tiny{$\pm$0.6}}         &     70.1\textsubscript{\tiny{$\pm$1.7}}     &  78.2\textsubscript{\tiny{$\pm$0.7}}      &   70.2\textsubscript{\tiny{$\pm$1.7}}    & 78.0\textsubscript{\tiny{$\pm$0.7}}       &       69.9\textsubscript{\tiny{$\pm$0.8}}   &   78.3\textsubscript{\tiny{$\pm$0.6}}    &   {71.3}\textsubscript{\tiny{$\pm$0.3}}      &  77.9\textsubscript{\tiny{$\pm$0.6}}    & 70.4\textsubscript{\tiny{$\pm$1.4}}  & \textbf{78.9}\textsubscript{\tiny{$\pm$0.5}} & \underline{72.8}\textsubscript{\tiny{$\pm$0.9}} \\
&    5\%        & 81.7\textsubscript{\tiny{$\pm$0.7}}        &     81.0\textsubscript{\tiny{$\pm$0.2}}    &  82.2\textsubscript{\tiny{$\pm$0.6}}         &     80.1\textsubscript{\tiny{$\pm$2.2}}     &   81.8\textsubscript{\tiny{$\pm$0.6}}    &  79.8\textsubscript{\tiny{$\pm$2.1}}     & 80.9\textsubscript{\tiny{$\pm$0.5}}        &     78.8\textsubscript{\tiny{$\pm$1.0}}    &   {82.4}\textsubscript{\tiny{$\pm$0.5}}       &   {81.8}\textsubscript{\tiny{$\pm$0.6}}     &  81.9\textsubscript{\tiny{$\pm$0.5}}    & 81.1\textsubscript{\tiny{$\pm$0.9}}  & \textbf{84.0}\textsubscript{\tiny{$\pm$1.4}} & \underline{83.9}\textsubscript{\tiny{$\pm$0.9}}\\
&    10\%       &  84.6\textsubscript{\tiny{$\pm$0.7}}         &   83.2\textsubscript{\tiny{$\pm$0.3}}     &  {85.6}\textsubscript{\tiny{$\pm$0.4}}       &   82.9\textsubscript{\tiny{$\pm$1.7}}       & 84.2\textsubscript{\tiny{$\pm$0.6}}       &  82.0\textsubscript{\tiny{$\pm$2.4}}     & 84.3\textsubscript{\tiny{$\pm$0.8}}        &    81.2\textsubscript{\tiny{$\pm$1.3}}      &  84.2\textsubscript{\tiny{$\pm$0.5}}     &    {83.6}\textsubscript{\tiny{$\pm$0.4}}    &  84.4\textsubscript{\tiny{$\pm$0.6}}  & 83.1\textsubscript{\tiny{$\pm$0.7}}  & \textbf{86.7}\textsubscript{\tiny{$\pm$1.5}} & \underline{86.4}\textsubscript{\tiny{$\pm$1.3}}\\

\midrule

\multirow{3}{*}{\textbf{Avg.}}
&   1\%         & 64.2 & 54.4 & 62.6 & 54.1 & 63.6 & 53.6 & 62.2 & 54.5 & 64.3 & {54.9} & {64.6} & 54.8 & \textbf{65.9}{{{\blue{$\uparrow$1.3}}}} & \underline{56.6}{{{\blue{$\uparrow$1.7}}}} \\
&   5\%         & 73.8 & 62.5 & 74.5 & 62.8 & 73.9 & 63.3 & 74.5 & 61.9 & 74.7 & {64.0} & {74.8} & {64.4} & \textbf{76.4}{{{\blue{$\uparrow$1.5}}}} & \underline{67.3}{{{\blue{$\uparrow$3.0}}}} \\
&   10\%        & 76.5 & 64.7 & {78.2} & 64.9 & 77.9 & 64.4 & 77.1 & 63.1 & 77.9 & 64.7 & 77.9 & {66.0} & \textbf{79.7}{{{\blue{$\uparrow$1.5}}}} & \underline{69.4}{{{\blue{$\uparrow$3.4}}}} \\

\bottomrule
\end{tabular}
}
\caption{In-distribution~(ID) and out-of-distribution~(OOD) accuracy of active learning methods across six datasets, evaluated at different percentages of the entire dataset size. Results are averaged over three runs with different random seeds. Bold indicates the best ID values, underlining marks the best OOD values, and values highlighted in blue show an improvement over the next best result.}
\label{tab:main_results}
\end{table*}

\paragraph{Implementation Details} We use the HuggingFace~\citep{wolf-etal-2020-transformers} implementation of BERT-base~\citep{devlin-etal-2019-bert} for our experiments. Following~\citet{margatina-etal-2021-active}, we set the annotation budget at 10\% of the unlabeled pool $\mathcal{U}$, initialize the labeled set at 0.1\% of $\mathcal{U}$, and the annotation batch size $b$ at 50. We train BERT-base models with a batch size of 16, learning rate of $2e-5$, using the AdamW~\citep{DBLP:conf/iclr/LoshchilovH19} optimizer with epsilon set to $1e-8$. For ALVIN, we set $K$ to 15, $\alpha$ to 2, and use the CLS token from the final layer to obtain representations and conduct interpolations. For other AL methods, we follow the same hyper-parameter tuning methods mentioned in their original papers. Each experiment is repeated three times with different random seeds, and we report the mean accuracy scores and standard deviations.

\section{Results}
\subsection{Main Results}
Table~\ref{tab:main_results} presents the main experimental results across the six datasets. Overall, we observe a considerable decline in OOD performance across all AL methods. ALVIN consistently outperforms all other AL methods in both in-distribution and out-of-distribution generalization. ALFA-Mix, CAL, and Uncertainty also show competitive performance, but do not surpass that of ALVIN. Notably, ALVIN enhances the effectiveness of Uncertainty, considerably improving performance compared to using Uncertainty alone. Finally, BADGE and BERT-KM demonstrate improvements only over Random sampling.

\begin{table}[ht!]
\centering
\small
\setlength{\tabcolsep}{4pt} % Slightly increased column spacing for better readability
\begin{tabular}{@{}cccccccc@{}} % Removed excess space around the table
\toprule
& \textbf{Method} & \textbf{AT} & \textbf{LN} & \textbf{NG} & \textbf{SE} & \textbf{WO} & \textbf{Avg.}\\
\midrule
\multirow{7}{*}{\rotatebox{90}{\textbf{NLI}}}
& Random & 13.8 & 43.7 & 37.5 & 44.4 & 45.4 & 37.0 \\
& Uncertainty & 12.2 & 49.9 & 39.6 & 47.6 & 48.1 & 39.5 \\
& BADGE & 16.2 & 50.5 & 43.3 & 49.2 & 48.0 & 41.4 \\
& BERT-KM & 10.6 & 46.6 & 39.1 & 47.0 & 47.4 & 38.1 \\
& CAL & 11.8 & 50.1 & 42.5 & 49.8 & 48.7 & 40.6 \\
& ALFA-Mix & 13.6 & 47.9 & 41.3 & 49.3 & 47.7 & 40.0 \\
& \textbf{ALVIN} & \textbf{18.2} & \textbf{54.1} & \textbf{48.3} & \textbf{52.8} & \textbf{53.6} & \textbf{45.4\textcolor{blue}{$\uparrow$4.0}} \\
\midrule
\multirow{7}{*}{\rotatebox{90}{\textbf{ANLI}}}
& Random & 83.2 & 29.9 & 31.4 & 29.7 & 41.7 & 43.2 \\
& Uncertainty & 85.0 & 32.5 & 30.7 & 29.8 & 41.8 & 44.0 \\
& BADGE & 62.4 & 30.2 & 33.3 & 30.2 & 39.5 & 39.1 \\
& BERT-KM & 74.3 & 28.6 & 30.2 & 29.4 & 37.4 & 40.0 \\
& CAL & 60.1 & 31.8 & 33.5 & 30.7 & 39.1 & 39.0 \\
& ALFA-Mix & 79.4 & 33.6 & 32.9 & 29.9 & 43.2 & 43.8 \\
& \textbf{ALVIN} & \textbf{85.8} & \textbf{42.2} & \textbf{40.2} & \textbf{39.8} & \textbf{50.5} & \textbf{51.7\textcolor{blue}{$\uparrow$7.7}} \\
\bottomrule
\end{tabular}
\caption{Out-of-distribution performance of active learning methods trained on NLI and ANLI datasets, evaluated using the NLI stress test. Values highlighted in blue indicate an improvement over the next best result.}
\label{tab:nli_stress}
\end{table}

\subsection{Additional OOD Generalization Results}
Following~\citet{deng-etal-2023-counterfactual}, we further evaluate the OOD generalization capabilities of models trained with various AL methods. Table~\ref{tab:nli_stress} presents the results on the NLI Stress Test \citep{naik-etal-2018-stress} for models trained on NLI and ANLI. We observe that ALVIN consistently outperforms all other AL methods in all stress tests, achieving an average performance improvement of 4.0 over BADGE, the next best performing method for models trained on NLI and, 7.7 over ALFA-Mix, the second best performing method for models trained on ANLI. Table~\ref{tab:amazon_ood} in the Appendix shows additional OOD results on Amazon reviews~\citep{ni-etal-2019-justifying}.

\section{Analysis}

\begin{table}[t!]
\centering
\begin{tabular}{lccc}
\toprule
\textbf{Method} & \textbf{Unc.} & \textbf{Div.}  & \textbf{Repr.}\\ 
\midrule
Random              & 0.121 & 0.641 & 0.584 \\  
Uncertainty         & 0.239 & 0.613 & 0.732 \\
BADGE               & 0.117 & 0.635 & 0.681 \\
BERT-KM             & 0.134 & 0.686 & 0.745 \\
CAL                 & 0.225 & 0.608 & 0.607 \\
ALFA-Mix            & 0.136 & 0.645 & 0.783 \\
\midrule
\textbf{ALVIN}               & 0.123 & {0.672} & {0.823} \\
\bottomrule
\end{tabular}
\caption{Uncertainty~(Unc.), diversity~(Div.), and representativeness~(Repr.) of unlabeled instances selected for annotation by active learning methods. Results are averaged across all datasets.}
\label{tab:instance_analysis}
\end{table}

\subsection{Characteristics of Selected Instances}
We analyze the characteristics of unlabeled instances identified through various active learning methods using uncertainty, diversity, and representativeness.

\paragraph{Uncertainty}
Following~\citet{yuan-etal-2020-cold}, we measure uncertainty with a model trained on the entire dataset to ensure that it provides reliable estimates. Specifically, we compute the average predictive entropy of the annotation batch $B$ via $-\frac{1}{|B|} \sum_{x \in B} \sum_{c=1}^C p(y=c|x) \log p(y=c|x)$, where $C$ is the number of classes.

\paragraph{Diversity}
We assess diversity in the representation space as proposed by~\citet{ein-dor-etal-2020-active}. For each instance $x_{i}$, diversity within the batch $B$ is calculated using $D(B) = \left(\frac{1}{|\mathcal{U}|} \sum_{x_i \in \mathcal{U}} \min_{x_j \in B} d(x_i, x_j)\right)^{-1}$, where $d(x_i, x_j)$ represents the Euclidean distance between $x_i$ and $x_j$.

\paragraph{Representativeness}
We measure the representativeness of instances in the annotation batch $B$, to ensure that the generated anchors do not attract outliers, which can negatively affect both in-distribution and out-of-distribution performance~\citep{karamcheti-etal-2021-mind}. To achieve this, we calculate the average Euclidean distance in the representation space between an example and its 10 most similar examples in $\mathcal{U}$, i.e., $R(x) = \frac{\sum_{x_i \in \text{KNN}(x)} \cos(x, x_i)}{K}$, where $\cos(x, x_i)$ is the cosine similarity between $x$ and its $k$-nearest neighbors, and $K$ is the number of nearest neighbors considered. Intuitively, a higher density degree within this neighborhood suggests that an instance is less likely to be an outlier~\citep{zhu-etal-2008-active, ein-dor-etal-2020-active}.

\begin{figure*}[ht!]
    \centering
    \begin{tikzpicture}
    \begin{groupplot}[
        group style={
            group size=3 by 1,  % Updated group size to 3 plots in 1 row
            horizontal sep=19pt,  % Adjusted space between plots to fit within margins
        },
        width=6cm,  % Adjusted width to accommodate better spacing and bar width
        height=6cm,  % Kept height consistent for a uniform look
        ymin=60, ymax=100,
        grid=major,
        grid style={dashed, gray!50},  % Lighter grid for less visual interference
        major grid style={line width=.2pt, draw=gray!50},
        tick label style={font=\footnotesize},
        label style={font=\small, font=\bfseries},
        minor x tick num=0,
        minor y tick num=0,
    ]
    % First Bar Chart, now with updated ALVIN performance values
    \nextgroupplot[
        ybar=0.5pt,  % Reduced overlap of bars
        enlarge x limits=0.2,  % Adjust spacing around bars
        ylabel={Accuracy (\%)},
        ylabel style={yshift=-1em},  % Bring ylabel closer to y-axis ticks
        symbolic x coords={ALVIN, ran, int-all, uni, kmean},
        xtick=data,
        xtick pos=bottom,
        ytick={70,80,90,100},
        ytick pos=left,
        bar width=10pt,  % Reduced bar width for better fit
        legend style={
            at={(0.98,0.98)}, anchor=north east, font=\small, inner sep=2pt,
            legend columns=1
        },
        cycle list={
            {fill=teal!70, draw=black}, % ID bars
            {fill=orange!70, draw=black}, % OOD bars
        },
    ]
    \addplot coordinates {(ALVIN,88.8) (ran,83.9) (int-all, 87.5) (uni, 84.5) (kmean, 87.1)}; % ID
    \addplot coordinates {(ALVIN,84.8) (ran,76.9) (int-all, 82.8) (uni, 78.2) (kmean, 83.6)}; % OOD
    \addlegendentry{ID}
    \addlegendentry{OOD}

    % Second Bar Chart, now with updated Beta(2, 2) performance values
    \nextgroupplot[
        ybar=0.5pt,  % Reduced overlap of bars
        enlarge x limits=0.8,  % Increased spacing around bars
        ylabel style={yshift=-1em},  % Bring ylabel closer to y-axis ticks
        symbolic x coords={Beta, Beta2},
        xtick={Beta, Beta2},
        xticklabels={0.5, 2},
        xtick pos=bottom,
        ytick={70,80,90,100},
        ytick pos=left,
        bar width=12pt,  % Adjusted bar width for better visualization
        legend style={
            at={(0.98,0.98)}, anchor=north east, font=\small, inner sep=2pt,
            legend columns=1
        },
        cycle list name=color list,
    ]
    \addplot[fill=blue!50] coordinates { (Beta,90.2) (Beta2,88.8)}; % ID
    \addplot[fill=red!60] coordinates { (Beta,82.4) (Beta2,84.8)}; % OOD
    \addlegendentry{ID}
    \addlegendentry{OOD}

    % Line Plot, now with updated K=15 performance values
    \nextgroupplot[
        % No ylabel for this plot
        xtick={1,5,10,15,20,25},
        xticklabels={1,5,10,15,20,25},
        ytick={60,70,80,90,100},
        xtick pos=bottom,
        ytick pos=left,
        xlabel={},  % Added x-axis label for the third plot
        cycle list name=color list,
        legend style={
            at={(0.98,0.98)}, anchor=north east,  % Top right for legend
            font=\small, inner sep=2pt,  % Match legend style and size
        },
    ]
    \addplot[blue, very thick, mark=*, mark options={solid, fill=blue!70, scale=1.5}] coordinates {
        (1, 80.8) (5, 83.7) (10, 86.8) (15, 88.8) (20, 88.3) (25, 87.1)
    };
    \addlegendentry{ID}

    \addplot[orange, very thick, dashed, mark=square*, mark options={solid, fill=orange!80, scale=1.5}] coordinates {
        (1, 77.8) (5, 80.9) (10, 83.0) (15, 84.8) (20, 83.9) (25, 82.2)
    };
    \addlegendentry{OOD}

    \end{groupplot}
    \end{tikzpicture}

    \vspace{-1em}
    \hspace{+1.9em}
    \subcaptionbox{ALVIN variants\label{fig:alvin_variant}}[0.3\textwidth][c]{\hspace{1em}}
    \hfill
    \subcaptionbox{Effect of $\alpha$\label{fig:lambda}}[0.3\textwidth][c]{}
    \hfill
    \subcaptionbox{Effect of $K$\label{fig:kappa}}[0.3\textwidth][c]{}
    \caption{Effects of different components of ALVIN and hyperparameter adjustments on both in-distribution (ID) and out-of-distribution (OOD) performance. Experiments are conducted on the IMDB dataset using 10\% of the acquired data.}
    \label{fig:alvin_plots}
\end{figure*}
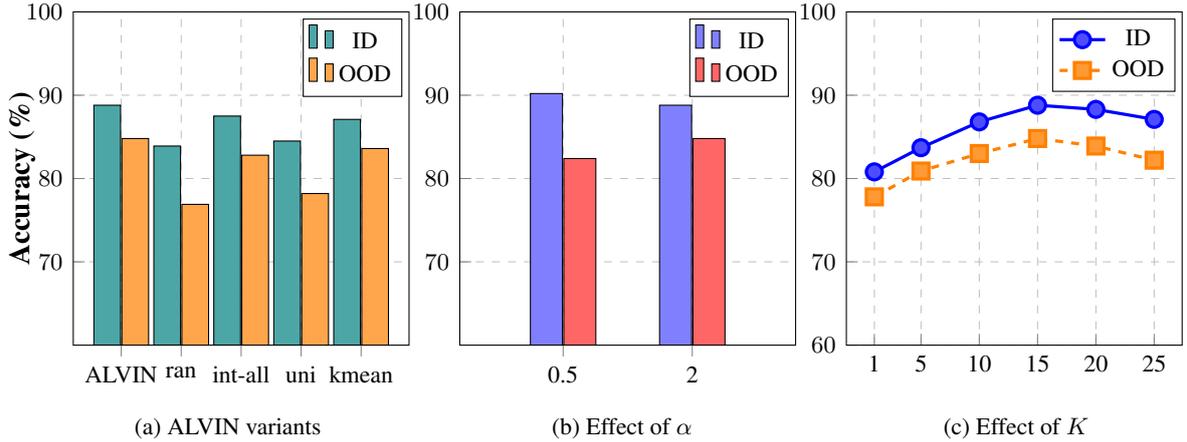

\paragraph{Results}
Table~\ref{tab:instance_analysis} presents the uncertainty, diversity, and representativeness metrics for unlabeled instances selected by different active learning methods. Uncertainty and CAL acquire the most uncertain examples, as indicated by their higher average entropy compared to other AL methods. Conversely, BADGE shows the lowest uncertainty, similar to ALFA-Mix and ALVIN.  BERT-KM scores highest in diversity, while Uncertainty exhibits the lowest score, suggesting that uncertainty sampling often selects similar examples near the decision boundary. Compared with other AL methods, ALVIN overall has a considerably better diversity. ALVIN achieves the highest representativeness score, indicating that its anchors are effectively positioned in the representation space to attract meaningful unlabeled instances without including outliers that could degrade model performance.

\begin{table}[]
\centering
\begin{tabular}{lccc}
\toprule
 & \textbf{1\%} & \textbf{5\%} & \textbf{10\%} \\
\midrule
NLI & 94.5 & 94.8 & 96.1 \\
ANLI & 93.6 & 94.2 & 95.4 \\
\bottomrule
\end{tabular}
\caption{Minority recall at different percentages of the dataset size.}
\label{tab:minority_recall}
\end{table}

\subsection{Effectiveness of Minority Identification}
To verify the reliability of using training dynamics for identifying minority examples, we validate the approach across different AL rounds. We calculate recall, defined as the fraction of ground-truth minority examples identified by our strategy. We conduct experiments on the NLI and ANLI datasets, where minority and majority examples are predefined. As shown in Table~\ref{tab:minority_recall}, relying on training dynamics provides consistent results, as the identified minority instances align with the ground-truth annotations.

\begin{table}[th!]
\centering
\setlength{\tabcolsep}{4pt}
\small
\begin{tabular}{l|cc|cc}
\toprule
& \multicolumn{2}{c|}{\textbf{Overlap}} & \multicolumn{2}{c}{\textbf{Negation}}\\
\textbf{Method} & Compr. $\downarrow$ & Acc. $\downarrow$ & Compr. $\downarrow$ & Acc. $\downarrow$ \\ 
\midrule
Random       & 3.6$\pm$0.5  & 85.8$\pm$0.5   & 3.8$\pm$0.6  & 87.2$\pm$1.7 \\ 
Uncertainty  & 3.3$\pm$0.4  & 85.2$\pm$1.2   & 4.3$\pm$0.2  & 93.7$\pm$1.8 \\
BADGE        & 3.5$\pm$0.2  & 86.2$\pm$0.9   & 4.1$\pm$0.5  & 93.2$\pm$1.6 \\
BERT-KM      & 3.1$\pm$0.6  & 84.5$\pm$0.5   & 3.9$\pm$0.3  & 91.5$\pm$2.2 \\
CAL          & 3.8$\pm$0.2  & 88.2$\pm$0.7   & 3.5$\pm$0.2  & 86.5$\pm$1.9 \\
Alfa-Mix     & 3.5$\pm$0.4  & 86.3$\pm$1.3   & 3.1$\pm$0.7  & 85.9$\pm$1.5 \\
\midrule
\textbf{ALVIN}      & \textbf{2.4}$\pm$\textbf{0.5}  & \textbf{80.7}$\pm$\textbf{0.8}   & \textbf{2.2}$\pm$\textbf{0.4}  & \textbf{82.6}$\pm$\textbf{1.8} \\
\bottomrule
\end{tabular}
\caption{Probing results for Overlap and Negation shortcut categories on the NLI dataset. Higher values in both compression~(Compr.) and accuracy~(Acc.) metrics indicate greater extractability of shortcut features from the model's representations.}
\label{tab:overlap_negation_results}
\end{table}

\subsection{Shortcut Extractability}
We evaluate the extractability of shortcut features from model representations using minimum description length probing~\citep{voita-titov-2020-information}. Our evaluation focuses on two common shortcuts: high-word overlap between the premise and hypothesis being labeled as ``entailment,'' and the presence of negation being labeled as ``contradiction.'' Higher probing accuracy and compression values suggest greater shortcut extractability. Table~\ref{tab:overlap_negation_results} presents the probing results on the NLI dataset for models trained with various AL methods over 10 rounds. We observe that prior AL methods increase shortcut extractability, as indicated by higher compression values and probing accuracies. In contrast, ALVIN exhibits the lowest compression values and probing accuracies.

\subsection{Hyperparameter Study}
We investigate the effect of the shape parameter $\alpha$ of Beta distribution on the overall performance of our proposed AL method. In Figure~\ref{fig:lambda} we present the performance of ALVIN when the\begin{enumerate*}[label=(\arabic*),before=\unskip{ Beta distribution }, itemjoin*={{, and }}]
  \item is U-shaped, i.e., $\alpha = 0.5$ 
  \item is bell-shaped, i.e., $\alpha = 2$.
  \end{enumerate*} When the distribution is U-shaped, this  leads to higher in-distribution accuracy but lower out-of-distribution generalization. This is due to the generated anchors being predominantly concentrated in two regions of the representation space, namely, those representing under-represented and well-represented groups. Due to the scarcity of examples in the under-represented group, anchors in this region fail to attract a sufficient number of instances, resulting in a tendency to attract examples from well-represented groups instead. Conversely, a bell-shaped distribution leads to anchors being dispersed across a wider range of the representation space, due to the broader variety of feature combinations it facilitates. Overall, adjusting the shape of the Beta distribution via $\alpha$ provides a means to balance the trade-off between in-distribution and out-of-distribution accuracy, potentially providing flexibility in the deployment of ALVIN depending on specific use-case requirements. Table~\ref{tab:asymetric_beta} in the Appendix presents additional results where the Beta distribution is asymmetric.

We also investigate the impact of the hyperparameter $K$, which determines the number of anchors generated between under-represented and well-represented example pairs. As illustrated in Figure~\ref{fig:kappa}, performance tends to be low with smaller $K$ values due to inadequate exploration of the representation space. However, as $K$ increases considerably, ALVIN's performance begins to align more closely with that of Uncertainty. This occurs because a larger number of anchors can cover a broader section of the representation space, thereby attracting high-uncertainty instances near the decision boundary.

\subsection{Ablations}
To better understand the effects of different components of ALVIN on both in-distribution and out-of-distribution performance, we conduct experiments with four ALVIN \begin{enumerate*}[label=(\arabic*), before=\unskip{ variants: }, itemjoin={{, }}]
    
    \item \textbf{ran} interpolates random pairs of labeled examples. The goal is to determine whether interpolations between under-represented and well-represented instances lead to the formation of meaningful anchors around unlabeled instances in the representation space

    \item \textbf{int-all} interpolates each minority example with every majority example, differing from the standard ALVIN practice which involves random pairings between under-represented and well-represented examples
    
    \item \textbf{uni} uniformly samples from $\mathcal{I}$~(line~\ref{algo:knn}) instead of using uncertainty to rank the unlabeled instances. It allows us to directly assess the impact of removing uncertainty-based selection
    
    \item \textbf{k-mean} clusters the samples from $\mathcal{I}$~(line~\ref{algo:knn}) via k-means, and then selects the unlabeled instances closest to the centroids of these clusters.
    
\end{enumerate*}

The results from Figure~\ref{fig:alvin_variant} demonstrate the performance of standard ALVIN is superior to that of its variants. Notably, interpolations between under-represented and well-represented examples considerably enhance performance, as evidenced by the drastic drop in performance observed with the \textbf{ran} variant. Interpolating between an under-represented example and all well-represented examples also leads to a slight reduction in performance. We hypothesize that this is due to the anchors being spread across a large area of the representation space, thus attracting repetitive high-uncertainty instances from well-represented groups. Additionally, integrating uncertainty into ALVIN helps refine the selection of unlabeled instances, providing a more informative subset for annotation. Finally, the \textbf{kmean} variant does not show improvement over standard ALVIN.

\begin{table}[t!]
\centering
\setlength{\tabcolsep}{12pt} % Increase padding
 \begin{tabular}{@{}lcc@{}}
 \toprule
  \textbf{Method} & \textbf{SST-2} & \textbf{IMDB} \\ 
 \midrule
Random & 0 & 0   \\  
Uncertainty  & 173  & 107  \\
BADGE  & 25640 & 3816  \\
BERT-KM  & 4265 & 431  \\
CAL & 708  & 273   \\
AlfaMix & 915 & 428  \\
\midrule
\textbf{ALVIN} & {781} & {357}  \\
\(\rightarrow\) \textit{Anchor Creation} & 468 & 232 \\
\(\rightarrow\) \textit{Example Selection} & 311 & 125 \\
 \bottomrule
  \end{tabular}
 \caption{Time taken~(in seconds) by active learning methods to select 100 instances from the unlabeled pool.}
\label{tab:runtime}
\end{table}

\subsection{Runtime}
We assess the computational runtime required for selecting instances for annotation, following the methodology of~\citet{margatina-etal-2021-active}. Specifically, we set the annotation batch size to 100, and conduct experiments using a Tesla V100 GPU. From Table~\ref{tab:runtime},  we see that Uncertainty is the most time-efficient AL method. Conversely, BADGE is the most computationally demanding AL method, as it involves clustering high-dimensional gradients. CAL ranks as the second most time-efficient method, followed by ALVIN, and ALFA-Mix. Overall, our approach demonstrates competitive speed compared to the fastest AL methods.

\section{Related Work}
\paragraph{Active Learning} AL methods can be categorized into three groups, informativeness-based, representativeness-based, and hybrid AL approaches~\citep{zhang-etal-2022-survey}. Informativeness-based AL approaches typically measure the usefulness of unlabeled instances via uncertainty sampling~\citep{DBLP:conf/sigir/LewisG94}, expected gradient length~\citep{DBLP:conf/nips/SettlesCR07}, and Bayesian methods~\citep{siddhant-lipton-2018-deep}. Recent AL works examine informativeness from the perspective of contrastive examples~\citep{margatina-etal-2021-active}, model training dynamics~\citep{zhang-plank-2021-cartography-active}, and adversarial perturbations~\citep{zhang-etal-2022-allsh}. Representativeness-based AL approaches like core-sets~\citep{DBLP:conf/iclr/SenerS18}, discriminative active learning~\citep{DBLP:journals/corr/abs-1907-06347}, and clustering-based methods~\citep{DBLP:journals/corr/abs-1901-05954, yu-etal-2022-actune} aim to select diverse instances such that the underlying task is better specified by the labeled set. Finally, hybrid AL approaches combine these two paradigms either by switching between informativeness and representantivess~\citep{DBLP:conf/aaai/HsuL15, fang-etal-2017-learning}, or by first creating informativeness-based representations of the unlabeled instances and then clustering them~\citep{DBLP:conf/iclr/AshZK0A20, ru-etal-2020-active}. Compared to prior work using interpolations for AL~\citep{DBLP:conf/cvpr/ParvanehATHHS22}, ALVIN differs in two key \begin{enumerate*}[label=(\arabic*), before=\unskip{ ways: }, itemjoin={{, and }}]

\item we opt for interpolations between specific labeled instance pairs, rather than randomly interpolating labeled and unlabeled instances

\item we sample $\lambda$ from a Beta distribution $\text{Beta}(\alpha, \alpha)$ instead of optimizing it for each pair individually. This approach  grants us greater control over the placement of the anchors in the representation space, ensuring they are positioned nearer to either under-represented or well-represented example groups as required.
\end{enumerate*}

\paragraph{Mixup} ALVIN is inspired by mixup~\citep{DBLP:conf/iclr/ZhangCDL18}, a popular data augmentation method originally explored in the field of computer vision. Mixup generates synthetic examples by interpolating random pairs of training examples and their labels. Recent mixup variants conduct interpolations using model representations~\citep{DBLP:conf/icml/VermaLBNMLB19}, dynamically compute the interpolation ratio~\citep{DBLP:conf/aaai/GuoMZ19, DBLP:journals/tnn/MaiHCSS22}, explore different interpolation strategies~\citep{yin-etal-2021-batchmixup}, and combine mixup with regularization techniques~\citep{jeong-etal-2022-augmenting, kong-etal-2022-dropmix}. In the context of NLP,~\citet{DBLP:journals/corr/abs-1905-08941} apply mixup on word and sentence embeddings using convolutional and recurrent neural networks. Conversely,~\citet{yoon-etal-2021-ssmix} propose a mixup variant that conducts interpolations on the input text.~\citet{park-caragea-2022-calibration} apply mixup to calibrate BERT and RoBERTa models, while~\citet{chen-etal-2020-mixtext} propose TMix, a mixup-inspired semi-supervised objective for text classification.

\section{Conclusion}
In this work, we propose ALVIN, an active learning method that uses intra-class interpolations between under-represented and well-represented examples to select instances for annotation. By doing so, ALVIN identifies informative unlabeled examples that expose the model to regions in the representation space which mitigate the effects of shortcut learning. Our experiments across six datasets, encompassing a broad range of NLP tasks, demonstrate that ALVIN consistently improves both in-distribution and out-of-distribution accuracy, outperforming other state-of-the-art active learning methods.

\section*{Limitations}
While we have demonstrated that ALVIN mitigates shortcut learning, we have not explored its ability to address fairness issues. ALVIN may inadvertently amplify biases present in the model's representations, as these are used to generate the anchors. Additionally, our experiments are limited to models trained with the masked language modeling pre-training objective, excluding other pre-training methods and model sizes. Finally, we acknowledge that active learning simulations are not always representative of real-world setups and annotation costs.

\section*{Acknowledgements}
Michalis Korakakis is supported by the Cambridge Commonwealth, European and International Trust, the ESRC Doctoral Training Partnership, and the Alan Turing Institute. Andreas Vlachos is supported by the ERC grant AVeriTeC (GA 865958). Adrian Weller acknowledges support from a Turing AI Fellowship under grant EP/V025279/1, and the Leverhulme Trust via CFI.

\bibliography{latex/anthology, latex/custom}

\appendix

\section{Appendix}
\subsection{Additional Results}\label{appendix:additional_ood}
\begin{table}[ht!]
\centering
\begin{tabular}{l c}
\toprule
\textbf{Method} & \textbf{Accuracy (\%)}\\
\midrule
Random & 86.56 \\
Uncertainty & 85.89 \\
BADGE & 83.23 \\
BERT-KM & 84.98 \\
CAL & 86.22 \\
ALFA-Mix & 86.18 \\
ALVIN & \textbf{89.75}\textcolor{blue}{$\uparrow$3.19} \\ % Improved performance
\bottomrule
\end{tabular}
\caption{Out-of-distribution performance of active learning methods trained on the SA dataset and evaluated on Amazon reviews. The value highlighted in blue indicates an improvement over the next best result.}
\label{tab:amazon_ood}
\end{table}
\begin{table}[ht!]
\centering
\begin{tabular}{ccc}
\toprule
\textbf{Beta} & \textbf{ID} & \textbf{OOD} \\
\midrule
$\alpha = 2, \beta = 5$ & 80.5 & 82.4 \\
$\alpha = 5, \beta = 2$ & 87.5 & 78.2 \\
$\alpha = 2, \beta = 2$ & 88.8 & 84.8 \\
\bottomrule
\end{tabular}
\caption{Comparison of ALVIN ID and OOD performance when Beta is asymmetric.}
\label{tab:asymetric_beta}
\end{table}

\end{document}